\pdfoutput=1

\documentclass[11pt]{article}

\usepackage[]{EMNLP2022}

\usepackage{times}
\usepackage{latexsym}

\usepackage[T1]{fontenc}

\usepackage[utf8]{inputenc}

\usepackage{microtype}

\usepackage{CJKutf8}

\usepackage{amsmath}
\usepackage{amssymb}

\usepackage{algorithm}
\usepackage{algpseudocode}

\usepackage{graphicx}
\usepackage{tabularx}
\usepackage{multirow}
\usepackage{adjustbox}
\usepackage{tikz}
\usepackage{pgfplots}
\usepackage{xcolor,colortbl}
\usepackage{array}

\newcommand*{\affaddr}[1]{#1} 
\newcommand*{\affmark}[1][*]{\textsuperscript{#1}}
\newcommand*{\email}[1]{\text{#1}}

\usepackage{inconsolata}

\title{ConNER: \\ Consistency Training for Cross-lingual Named Entity Recognition}

\author{Ran Zhou\thanks{~~Ran Zhou is under the Joint Ph.D. Program between Alibaba and Nanyang Technological University.}\affmark[~~1,2]~~~Xin Li\affmark[1]\thanks{~~Corresponding author}~~~~Lidong Bing\affmark[1]~~~Erik Cambria\affmark[2]~~~Luo Si\affmark[1]~~~Chunyan Miao\affmark[2]\\
	\affaddr{\affmark[1]DAMO Academy, Alibaba Group}\quad
	\affaddr{\affmark[2]Nanyang Technological University, Singapore}\\
	\email{{\tt\{ran.zhou, xinting.lx, l.bing, luo.si\}@alibaba-inc.com}}\\
	\email{{\tt\{cambria, ascymiao\}@ntu.edu.sg}} \\
	}

\begin{document}
\maketitle
\begin{abstract}

Cross-lingual named entity recognition (NER) suffers from data scarcity in the target languages, especially under zero-shot settings. 
Existing translate-train or knowledge distillation methods attempt to bridge the language gap, but often introduce a high level of noise. To solve this problem, consistency training methods regularize the model to be robust towards perturbations on data or hidden states.
However, such methods are likely to violate the consistency hypothesis, or mainly focus on coarse-grain consistency.
We propose ConNER as a novel consistency training framework for cross-lingual NER, which comprises of: (1) translation-based consistency training on unlabeled target-language data, and (2) dropout-based consistency training on labeled source-language data. 
ConNER effectively leverages unlabeled target-language data and alleviates overfitting on the source language to enhance the cross-lingual adaptability. 
Experimental results show our ConNER achieves consistent improvement over various baseline methods.\footnote{Our code is available at \url{https://github.com/RandyZhouRan/ConNER}.}

\end{abstract}

\section{Introduction}

With the emergence of large scale multilingual pretrained language models~\citep{devlin2019bert,conneau2020unsupervised}, cross-lingual named entity recognition (NER) performance has seen substantial improvement, especially under zero-shot setting where no labeled target-language training data is available~\citep{tsai2016cross,xie2018neural,jain-etal-2019-entity}. 
Nevertheless, due to linguistic gaps between languages, NER models trained solely on the source-language data are likely to overfit the source language's characteristics, and still suffer from sub-optimal performance when tested on target languages directly. 

To further enhance cross-lingual adaptability, various methods have been proposed to explicitly incorporate target-language information during training. These approaches can be roughly categorized into: (1) translate-train and (2) knowledge distillation. 
Translate-train produces parallel target-language training data by translating the source-language training data and mapping the labels~\citep{xie2018neural,bari2020zero,li2020unsupervised}. However, the translated sentences are less natural and the label mapping is also error-prone. Moreover, without utilizing the abundant unlabeled target-language data, their performance is usually sub-optimal~\citep{ijcai2020-543}.
On the other hand, knowledge distillation trains the student model on target-language data with soft labels from a teacher model~\citep{wu-etal-2020-single,ijcai2020-543,chen-etal-2021-advpicker,liang2021reinforced,zhodoe}. Given the relatively low performance of the teacher model on target languages, the soft labels are noisy and limit the efficacy of unlabeled target-language data when using these labels as supervision signals.

In this paper, we primarily explore how to better utilize unlabeled target-language data.
Instead of adopting knowledge distillation, which is vulnerable to the noise in soft labels, we adopt consistency training as a more fault-tolerant strategy. In general, consistency training works by enhancing the smoothness of output distributions and improving the model's robustness and generalization~\citep{miyato2018virtual}. Its training process is less sensitive towards noisy labels by regulating multiple predictions on different views of the same data point. 
Several works already attempted to apply consistency training on NER, including token-level and sequence-level consistency methods. 
Token-level consistency works on the same granularity as NER by regularizing the model to be invariant towards Gaussian noise~\citep{zheng2021consistency} or word replacement~\citep{lowell2020unsupervised}. \\

However, such na\"ive noise or augmentation might violate the assumption that noised tokens share the same labels as the original ones.
On the other hand, a sequence-level consistency method~\citep{wang2021unsupervised} attempted to adapt back-translation-based~\citep{edunov2018understanding} consistency to NER. Due to difficulties in word alignment, they only use a coarse-grain consistency on the appearance of a certain type of entity in the original and back-translated sentences, which is plausibly a sub-optimal design for token-level tasks like NER. To this end, other works used constituent-based tagging schemes~\cite{zhoext}.

We propose ConNER, a cross-lingual NER framework that primarily leverages consistency training on unlabeled target-language data using translation.
Concretely, ConNER encourages the model to output consistent predictions between a span of tokens in the original sentence and their projection in the translated sentence. 
We tackle the problem of word alignment with alignment-free translation, and propose span-level consistency constraints to overcome the problem of token number changes during translation.
Another advantage of translating unlabeled target-language data into the source language is that we can obtain more reliable training signals from the predictions on these translated sentences, because the model is trained with supervision from labeled source-language data and has better performance on the source language.
Such reliable signals can be propagated back to the unlabeled target-language sentences through regularizing the prediction consistency between aligned spans.
Furthermore, consistency training on the parallel unlabeled data helps to align different languages in the same representation space for better cross-lingual adaptability.

To mitigate overfitting on the source language, we also introduce a dropout-based consistency training on labeled source-language data, where we train our model to be robust towards noise induced by different dropout processes.
In summary, ConNER enhances the model's robustness towards both translation-based and dropout-based perturbations, which alleviates overfitting and achieves better cross-lingual generalization.
To illustrate the effectiveness of ConNER, we conduct experiments on three transfer pairs and benchmark against various baseline methods. Experimental results show that ConNER achieves substantial improvement on cross-lingual NER over the baselines.


Our major contributions are as follows: (1) We propose a novel consistency training framework for cross-lingual NER and achieve consistent improvement over multiple baselines across various transfer pairs. (2) We present translation-based consistency training to effectively utilize unlabeled target-language data. It obviates using word-alignment tools and handles token number changes. (3) We introduce dropout-based consistency training to NER and reinforce the model's robustness towards input noise and alleviate overfitting on labeled source-language data.


\section{Methodology}

\begin{figure*}
    \centering
    \includegraphics[width=\textwidth]{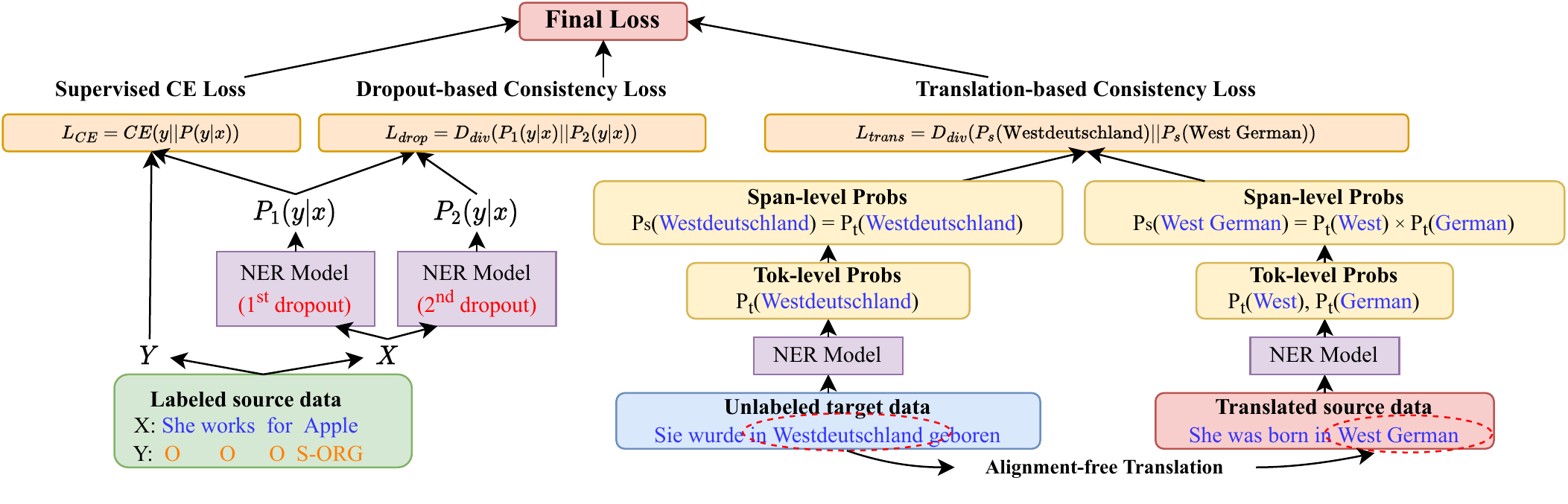}
    \caption{ConNER: consistency training for cross-lingual NER.}
    \label{fig:conner}
\end{figure*}

In this section, we introduce ConNER, our consistency training framework for cross-lingual NER. 
As shown in Fig.~\ref{fig:conner}, the overall framework consists of three components, namely (1) training with supervised cross-entropy loss, (2) dropout-based consistency training and (3) translation-based consistency training.
Specifically, supervised cross-entropy loss is calculated on labeled source-language data, similar to a vanilla NER model. 
Secondly, dropout-out based consistency training (Section~\ref{sec:dropout}) feeds the same labeled source-language sample twice through the model, and enforces the model to output consistent probability distributions on the same token from two different dropout operations.
Lastly, translation-based consistency training (Section~\ref{sec:trans}) translates unlabeled target-language data into the source language via alignment-free translation method, and encourages the model to make consistent predictions on conjugate spans between the original and translated sentences.

\subsection{Problem Definition}
\label{sec:definition}

Following previous works, we formulate the task of cross-lingual NER as a sequence tagging problem. Given an input sentence comprising of $n$ tokens $\mathbf{X}=\{x_{1}, x_{2}, ... , x_{n}\}$, 
we aim to assign an NER tag $y_{i} \in \mathcal{Y}$ to each token $x_{i} \in \mathbf{X}$, where $\mathcal{Y}$ denotes the set of all possible NER tags. 
When there are a total number of $N$ entity types $\mathcal{C}=\{C_{1},C_{2},...,C_{N}\}$ with BIOES scheme $\mathcal{P}=\{\textrm{ B-, I-, E-, S-}\}$, the token-level label space is $\mathcal{Y}= \mathcal{P} \times \mathcal{C} \cup \{O\}$. 
In our framework, apart from labeled source-language training data $\mathcal{D}_{l}^{src}$, unlabeled target-language data $\mathcal{D}_{u}^{tgt}$ is also available during training.

\subsection{Consistency Training for NER}

Consistency training~\citep{miyato2018virtual,clark-etal-2018-semi,xie2020unsupervised} aims to enhance the model's robustness by regularizing model predictions to be invariant to label-preserving perturbations. Despite its success on CV and sentence-level NLP tasks~\citep{miyato2018virtual,xie2020unsupervised}, consistency training for NER remains largely underexplored, where the model should be consistent on token-level or entity-level predictions.


In this work, we explore better consistency training methods for cross-lingual NER, which can be formulated as follows.
Let $\mathbf{\phi}$ be a transformation function for generating small perturbation, such as text translation or hidden state noising. Given an input sequence of tokens (i.e., a sentence $\mathbf{X}$) or hidden states, and a segment $s$ in the sequence (e.g., a span of tokens), we apply $\mathbf{\phi}$ on the input to obtain a transformed sequence which contains the transformation of $s$, denoted as $s'$. For example, we translate $\mathbf{X}$ to another language and $s'$ is the translation of $s$ in the translated sentence. We refer to $(s,s')$ as a pair of conjugate segments. We encourage the model to output consistent probability distributions on $s$ and $s'$, by minimizing the consistency loss as:
\begin{equation}
\label{eqn:L_c}
L_{c}=\frac{1}{m}\sum_{\substack{s\in\mathbf{X}, \\ s'\in\phi(\mathbf{X})}}D_{div}(P(y~|~s)~||~P(y~|~s'))
\end{equation}
where $D_{div}$ is a measure of divergence, $m$ is the total number of conjugate segment pairs. 

In this work, we introduce two variants of consistency training for NER which vary in the form of  perturbation $\mathbf{\phi}$: (1) translation-based perturbation on input text, and (2) dropout-based perturbation on token representations. 

Depending on the form of perturbation, $s$ can be a single token or a span of tokens, and correspondingly $P(y|s)$ will be token-level or span-level probability. We explain in details in the following Sec.~\ref{sec:trans} and~\ref{sec:dropout}.

\subsection{Translation-based Consistency Training}
\label{sec:trans}

In this section, we present translation-based consistency training as the first variant, where we introduce a textual-level perturbation $\mathbf{\phi}$ by translating the original sentence into a different language. 
The conjugate segments $(s, s')$ are now a span of tokens in the original sentence and its translation in the translated sentence, respectively. By enforcing the model to output consistent predictions on $s$ and $s'$, the model tends to capture language-invariant features and thus has better cross-lingual adaptability.

To enable the above translation-based consistency training for NER, there are still two challenging issues to be resolved: (1) the word alignments between the original and translated sentences are not available and the existing alignment tools for inducing word alignments are often error-prone; (2) the perturbed sentence (i.e., the translation) may have different length and word order, and it is likely that the token-level one-to-one correspondence for calculating consistency loss will not be applicable.
To tackle these challenges, we propose a novel method which leverages alignment-free translation to automatically construct span-level correspondence, and caters for token number changes as well. Note that our approach is not sensitive to the change of sequence length after perturbation because it no longer requires the consistency on token-level predictions.

\subsubsection*{Alignment-free Translation}

Given an input sentence $\mathbf{X}=\{x_{1},...,x_{n}\}$ which contains a candidate span $s_{ij}=\{x_{i},...,x_{j}\}$, we aim to translate it into a different language as $\mathbf{X'}=\{x'_{1},...,x'_{n'}\}$ and locate the span $s'_{kl}=\{x'_{k},...,x'_{l}\}$ that corresponds to $s_{ij}$. 
Adapting the method proposed in~\citet{liu2021mulda},
we conduct alignment-free translation to automatically locate $s'_{kl}$ in the translated sentence. Specifically, as shown in Fig.~\ref{fig:translation}, we first replace the candidate span $s_{ij}$ with a placeholder token SPAN\textsubscript{ij} to obtain $\hat{\mathbf{X}}=\{x_{1},...,x_{i-1},\textrm{SPAN}\textsubscript{ij},x_{j+1},...,x_{n}\}$. Then, we feed $\hat{\mathbf{X}}$ through a translation engine to obtain its translation $\hat{\mathbf{X'}}$. We notice the placeholder SPAN\textsubscript{ij} remains mostly unchanged in the translated sentence, and it indicates where the candidate span should appear in the translation. 
Next, we aim to revert the placeholder back to real text, by replacing it with the translation of $s_{ij}$, denoted as $s'_{kl}$. Lastly, we substitute the placeholder in $\hat{\mathbf{X'}}$ with $s'_{kl}$, to obtain the final translation $\mathbf{X'}$, and $(s_{ij},s'_{kl})$ is recorded as a pair of conjugate segments.

\begin{figure}[t]
    \centering
    \adjustbox{minipage=[r][11.5cm][b]{0.6\textwidth},scale={0.85}}{
    \textbf{Input sentence with candidate span marked in blue:} \\
    \hspace*{0.4cm} Bruce Willis wurde in \textcolor{blue}{Westdeutschland} geboren. \\
    
    \textbf{1. Replace span with placeholder and translate:} \\
    \hspace*{0.4cm} Bruce Willis wurde in \textcolor{orange}{SPAN\textsubscript{44}} geboren. \\
    $\Rightarrow$ Bruce Willis was born in \textcolor{orange}{SPAN\textsubscript{44}}. \\
    
    \textbf{2. Obtain translation of span:} \\
    \hspace*{0.4cm} \textcolor{blue}{Westdeutschland} $\Rightarrow$ \textcolor{blue}{West German} \\
    
    
    \textbf{3. Replace placeholder with translated span \\ \hspace*{0.3cm} and extract conjugate pair:} \\
    \hspace*{0.4cm} Bruce Willis was born in \textcolor{blue}{West German}. \\
    \hspace*{0.4cm} \textbf{Conjugate pair:} \textcolor{blue}{(Westdeutschland, West German)} \\
    
    \textbf{4. Converting token-level probs to span-level probs:} \\
    \small
    $P_{s}(\textrm{PER}|\textit{West German})=P_{t}(\textrm{B-PER}|\textit{West}) \cdot P_{t}(\textrm{E-PER}|\textit{German})$ \\
    $P_{s}(\textrm{LOC}|\textit{West German})=P_{t}(\textrm{B-LOC}|\textit{West}) \cdot P_{t}(\textrm{E-LOC}|\textit{German})$ \\
    $P_{s}(\textrm{ORG}|\textit{West German})=P_{t}(\textrm{B-ORG}|\textit{West}) \cdot P_{t}(\textrm{E-ORG}|\textit{German})$ \\
    $P_{s}(\textrm{MISC}|\textit{West German})=P_{t}(\textrm{B-MISC}|\textit{West}) \cdot P_{t}(\textrm{E-MISC}|\textit{German})$ \\
    $P_{s}(\textrm{O}|\textit{West German})=P_{t}(\textrm{O}|\textit{West}) \cdot P_{t}(\textrm{O}|\textit{German})$ \\
    $P_{s}(\textit{illegal}|\textit{West German})=1 - P_{s}(\textrm{PER}) - P_{s}(\textrm{LOC}) \\ 
    \hspace*{3.5cm} - P_{s}(\textrm{ORG}) - P_{s}(\textrm{MISC}) - P_{s}(\textrm{O})$ \\
    }
    \caption{Example illustrating the key steps of translation-based consistency training.}
    \label{fig:translation}
\end{figure}

\subsubsection*{Token-level to Span-level Probability}

As mentioned above, $s_{ij}$ and $s'_{kl}$ might have different number of tokens, and thus it is intractable to calculate one-to-one consistency loss between pairs of single tokens. 
We resolve this issue by converting token-level probability distributions (over label space $\mathcal{Y}$ defined in Sec.~\ref{sec:definition}) to span-level probability distributions over the set of possible entity types, i.e., $\mathcal{C} \cup \{O\}$. 
Specifically, given the last hidden state $h_{i}$ of token $x_{i}$, the NER model feeds it through a linear layer to obtain token-level probability distribution $P_{t}(y_{i}|x_{i},h_{i})$. 
As shown in Algorithm~\ref{alg:convert}, to obtain span-level probability distribution $P_{s}(z_{ij}|s_{ij})$, we multiply probabilities of token-level labels which form a legal label sequence under BIOES scheme. Step 4 in Fig.~\ref{fig:translation} also gives an example calculation as illustration.

\begin{algorithm}[t]
\small
\caption{Token to span probability conversion}
\label{alg:convert}
\begin{algorithmic}
\State Given candidate span $s_{ij}=\{x_{i},...x_{j}\}$ and token-level probabilities $\{P_{t}(y_{u}|x_{u})~|~u\in[i,j]\}$
\State $P_{s}(\textrm{O}|s_{ij}) \gets 1$, $P_{s}(\textit{illegal}|s_{ij}) \gets 1$
\For{$\textrm{CLS} \in \mathcal{C}$}
    \If{$i=j$} 
        \State $P_{s}(\textrm{CLS}|s_{ij}) \gets P_{t}(\textrm{S-CLS}|x_{i})$
    \ElsIf{$j-i=1$} 
        \State $P_{s}(\textrm{CLS}|s_{ij}) \gets P_{t}(\textrm{B-CLS}|x_{i}) \times P_{t}(\textrm{E-CLS}|x_{i})$
    \Else
        \State $P_{s}(\textrm{CLS}|s_{ij}) \gets P_{t}(\textrm{B-CLS}|x_{i})$
        \For{$u\in(i,j)$}
            \State $P_{s}(\textrm{CLS}|s_{ij}) \gets P_{s}(\textrm{CLS}|s_{ij}) \times P_{t}(\textrm{I-CLS}|x_{u})$
        \EndFor
        \State $P_{s}(\textrm{CLS}|s_{ij}) \gets P_{s}(\textrm{CLS}|s_{ij}) \times P_{t}(\textrm{E-CLS}|x_{j})$
    \EndIf 
    \State $P_{s}(\textit{illegal}|s_{ij}) \gets P_{s}(\textit{illegal}|s_{ij})-P_{s}(\textrm{CLS}|s_{ij})$    
\EndFor

\For{$u\in[i,j]$}
    \State $P_{s}(\textrm{O}|s_{ij}) \gets P_{s}(\textrm{O}|s_{ij}) \times P_{t}(\textrm{O}|x_{u})$
\EndFor
\State $P_{s}(\textit{illegal}|s_{ij}) \gets P_{s}(\textit{illegal}|s_{ij})-P_{s}(\textrm{O}|s_{ij})$   

\State $P_{s}(z_{ij}|s_{ij})=\{P_{s}(\textrm{CLS\textsubscript{1}}|s_{ij}),...,P_{s}(\textrm{CLS\textsubscript{N}}|s_{ij}), \newline \hspace*{5em} P_{s}(\textrm{O}|s_{ij}), P_{s}(\textit{illegal}|s_{ij})\}$
\State \textbf{return} $P_{s}(z_{ij}|s_{ij})$

\end{algorithmic}
\end{algorithm}

Note that we introduce an extra span-level class \texttt{illegal} to include all label sequences of $s_{ij}$ that violate the BIOES rules (e.g., \{S-PER, I-PER\}), such that all entries of $P_{s}(z_{ij}|s_{ij})$ add up to one to form a valid probability distribution. In other words, the label space of span-level label $z_{ij}$ is $\mathcal{C} \cup \{\textrm{O}, \texttt{illegal}\}$.
We apply the same transformation on $s'_{kl}$ to obtain its span-level distribution $P_{s}(s'_{kl})$

Finally, we calculate $D_{div}$ between span-level probabilities of each conjugate segment pair $(s_{ij}, s'_{kl})$ using bidirectional Kullback-Leibler (KL) divergence, and Eqn.~\ref{eqn:L_c} becomes:
\begin{equation}
\small
\begin{aligned}
L_{\textrm{trans}}=\frac{1}{m}\sum_{\substack{s_{ij}\in\mathbf{X} \\ s'_{kl}\in\mathbf{X'}}}  \frac{1}{2}~[~&\textrm{KL}(P_{s}(z_{ij}|s_{ij})~||~P_{s}(z_{kl}|s'_{kl}))~ \\
+~&\textrm{KL}(P_{s}(z_{kl}|s'_{kl})~||~P_{s}(z_{ij}|s_{ij}))]
\end{aligned}
\end{equation}

$ $

In our experiments, we apply translation-based consistency training on unlabeled target-language sentences by translating them into the source language. 
It is noteworthy that, although our translation-based consistency training is applicable to arbitrary span pairs from the parallel sentences, it would be more valuable for NER task if the selected spans are likely to be entities.
Therefore, we use a weak NER tagger trained with only labeled source-language data to assign NER labels on unlabeled target-language sentences, and use the predicted entities as our candidate spans for consistency training. 
Note that we only use predictions from the weak tagger to locate the boundary of candidate spans, which to some extent reduces the impact of predictions with wrong entity types.

\subsection{Dropout-based Consistency Training}
\label{sec:dropout}

We introduce another variant of consistency training that is based on dropout~\citep{srivastava2014dropout}. Inspired by~\citet{wu2021r,gao-etal-2021-simcse}, we consider dropout as a form of perturbation $\mathbf{\phi}$ at representational level. By regularizing the model to be invariant to different random dropouts, we encourage the model to make predictions based on more diverse features rather than overfitting on certain spurious features. As word order does not change after applying dropout, the conjugate segments now become the representations of the same token undergoing different dropout processes.

Concretely, we pass the same sentence $\mathbf{X}$ through the encoder twice. As a result of different stochastic dropout in each pass, we obtain two different sets of token representations for $\mathbf{X}$.
Subsequently, the model outputs two different token-level probability distributions $P_{1}(y_{i}|x_{i})$ and $P_{2}(y_{i}|x_{i})$ over the label space $\mathcal{Y}$.
We also adopt bidirectional KL divergence as $D_{div}$, and calculate dropout-based consistency loss as:
\begin{equation}
\small
\begin{aligned}
L_{\textrm{drop}}   =\frac{1}{n}\sum_{x_{i}\in\mathbf{X}}  \frac{1}{2}~[~&\textrm{KL}(P_{1}(y_{i}|x_{i})~||~P_{2}(y_{i}|x_{i})~ \\
        +~&\textrm{KL}(P_{2}(y_{i}|x_{i})~||~P_{1}(y_{i}|x_{i}))]  
\end{aligned}
\end{equation}

We apply dropout-based consistency training on labeled source-language data in our experiments, but NER labels are not used.

\subsection{Training Objective}

As mentioned above, we apply translation-based consistency training on unlabeled target-language data, and dropout-based consistency training on labeled source-language data. 

These consistency losses are combined with the supervised cross-entropy loss (i.e., $L_{\textrm{CE}}$) on labeled source-language data, which gives the total training objective as:
\begin{equation}
\small
\begin{aligned}
L_{\textrm{total}}  = \sum_{X\in\mathcal{D}_{l}^{src}}(L_{\textrm{CE}} + \alpha~L_{\textrm{drop}}) +\sum_{X\in\mathcal{D}_{u}^{tgt}}\beta~L_{\textrm{trans}}   
\end{aligned}
\end{equation}
where $\alpha$ and $\beta$ are weight coefficients. 

\section{Experiments}

In this section, we evaluate our consistency training framework ConNER on cross-lingual NER, and compare with various state-of-the-art models.

\subsection{Dataset}

We conduct experiments on CoNLL02 and CoNLL03 datasets~\citep{tjong-kim-sang-2002-introduction, sang2003introduction}, which involve four languages: English (En), German (De), Spanish (Es) and Dutch (Nl). 
To evaluate on more distant transfer pairs, we also experiment on WikiAnn dataset~\citep{pan-etal-2017-cross} of English (En), Chinese (Zh), Arabic (Ar) and Hindi (Hi).
We adopt the BIOES entity annotation scheme in our experiments. 
Following previous zero-shot cross-lingual NER works, we use the original English training set as our training data $\mathcal{D}_{l}^{src}$, while treating all other languages as target languages and evaluate on their test sets. We also use the original English development set for early-stopping and model selection. The NER labels from target language training sets are removed, and they are used as our unlabeled target-language data $\mathcal{D}_{u}^{tgt}$.

\subsection{Implementation Details}

We implement our vanilla NER model using XLM-RoBERTa-large~\citep{conneau2020unsupervised} with CRF head~\citep{lample2016neural}. The emission probabilities of the CRF layer, which are obtained by feeding the last hidden states of XLM-R through a feed-forward layer, are used as the token-level probabilities for calculating consistency losses in Section~\ref{sec:trans} and~\ref{sec:dropout}. We use AdamW optimizer~\citep{loshchilov2018decoupled} with learning rate $l_{r}=2\mathrm{e}{-5}$ and set batch sizes for both labeled and unlabeled data as 16. We train the NER model for 10 epochs and select the best checkpoint using English dev set. The model is evaluated on target-language test sets and we report the averaged micro-F1 score over 3 runs. 
We use Google Translate as our translation engine for the main experiments.

\begin{figure}[t!]
\begin{tikzpicture}
\pgfplotsset{width=8cm,height=5cm,compat=newest}
\begin{axis}[
    xtick={0.25,0.5,1.0,2.0,4.0},
    ymin=96.4, ymax=96.8,
    symbolic x coords={0.25,0.5,1.0,2.0,4.0},
    xticklabel style = {font=\fontsize{6}{1}\selectfont},
    yticklabel style = {font=\fontsize{6}{1}\selectfont},
    legend style={font=\fontsize{5}{1}\selectfont},
  ylabel={\footnotesize Avg dev F$_1$},
  xlabel={\footnotesize $\alpha$},
  enlargelimits=0.1,
  legend style={at={(0.35,0.1)},anchor=south,legend columns=2}, 
  every axis plot/.append style={thick},
  tick label style={/pgf/number format/fixed},
    every node near coord/.append style={font=\tiny}
]
 
\addplot[green!60!black] [mark=square*]  coordinates {(0.25, 96.6730) (0.5, 96.5830) (1.0, 96.5897) (2.0, 96.7030) (4.0, 96.5663)};
\addplot[blue!70!white] [mark=diamond*]  coordinates {(0.25,96.7163)	(0.5,96.7500)	(1.0,96.7000)	(2.0,96.6663)	(4.0,96.4163)}; 
\addplot[orange] [mark=triangle*]  coordinates {(0.25,96.7030)	(0.5,96.5763)	(1.0,96.6463)	(2.0,96.6197)	(4.0,96.6330)}; 
\addplot[red!70!white] [mark=otimes*]  coordinates {(0.25,96.7230)	(0.5,96.6530)	(1.0,	96.7063)	(2.0,96.6497)	(4.0,96.5430)}; 
\addplot[cyan] [mark=x]  coordinates {(0.25,96.5700)	(0.5,96.6300)	(1.0,	96.6000)	(2.0,96.5963)	(4.0,96.5195)}; 

\addlegendentry{$\beta=0.25$}
\addlegendentry{$\beta=0.5$}
\addlegendentry{$\beta=1.0$}
\addlegendentry{$\beta=2.0$}
\addlegendentry{$\beta=4.0$}

\end{axis}
\end{tikzpicture}

\vspace{0.1cm}
\caption{Averaged dev F1 for hyperparameter tuning.}
\label{fig:hyperparameter}
\end{figure}
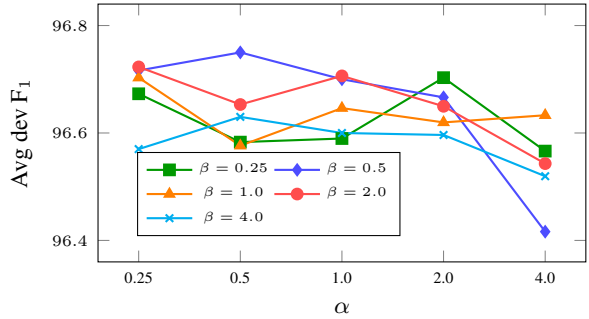

We conduct a grid search on both hyperparameters $\alpha$ and $\beta$ over the range of $\{0.25,0.5,1.0,2.0,4.0\}$. The English dev set F1 averaged across three transfer pairs are shown in Fig.~\ref{fig:hyperparameter}. We set $\alpha=0.5, \beta=0.5$ based on the best dev set performance.

\begin{table*}[h!]
    \centering
    \begin{tabular}{p{7cm}cccc}
    \hline\hline
         \textbf{Method} & \textbf{De} & \textbf{Es} & \textbf{Nl} & \textbf{Avg} \\
         \hline
         \hspace{3mm} vanilla baseline & 72.10	& 79.64 & 81.29 & 77.68\\
         \hline
         \multicolumn{5}{l}{\textit{Translate-train}} \\
         \hspace{3mm} BWET~\citep{xie2018neural}\textsuperscript{\dag} & 73.84 &	79.03 &	80.93 & 77.93\\
         \hspace{3mm} MTL-WS~\citep{li2020unsupervised} & 74.92 & 76.73 & 78.62 & 76.76\\
         \hspace{3mm} MulDA~\citep{liu2021mulda}\textsuperscript{\ddag} & 74.55 & 78.14 & 80.22 & 77.64\\
         \hline
         \multicolumn{5}{l}{\textit{Knowledge-distillation}} \\
         \hspace{3mm} TSL~\citep{wu-etal-2020-single} & 73.16 & 76.75 & 80.44 & 76.78\\
         \hspace{3mm} Unitrans~\citep{ijcai2020-543} & 74.82 & 79.31 & 82.90 & 79.01\\
         \hspace{3mm} AdvPicker~\citep{chen-etal-2021-advpicker} & 75.01 & 79.00 & 82.90 & 78.97\\
         \hline
         \multicolumn{5}{l}{\textit{Consistency training}} \\
         \hspace{3mm} MLM-replace~\citep{lowell2020unsupervised}\textsuperscript{\ddag} & 71.65 & 79.70 & 80.83 & 77.39\\
         \hspace{3mm} xTune~\citep{zheng2021consistency}\textsuperscript{\ddag} & 74.78 & 80.03 & 81.76 & 78.85\\
         \hspace{3mm} \textbf{ConNER} \textit{(Ours)} & \textbf{77.14}	& \textbf{80.50}	& \textbf{83.23}	& \textbf{80.29}\\
         \hline
         \rowcolor{black!10!white}
         \multicolumn{5}{l}{\textit{Multi-task}} \\
         \rowcolor{black!10!white}
         \hspace{3mm} TOF~\citep{zhang2021target}\textsuperscript{*} & 76.57 & 80.35 &  82.79 & 79.90\\
         
    \hline\hline
         
    \end{tabular}
    \caption{Experimental results of cross-lingual NER methods. \textsuperscript{\dag} denotes reproduced results with XLM-R-large, which are higher than the original paper based on LSTM. \textsuperscript{\ddag} denotes reimplementation on our datasets. Results without markers are cited from the original papers. \textsuperscript{*}\citet{zhang2021target} use extra labeled MRC data for multi-task training, thus it is not directly comparable with our method.}
    \label{tab:main_result}
\end{table*}

\subsection{Main Results}

As shown in Table~\ref{tab:main_result}, ConNER achieves significant performance improvement (avg. 2.61) over the zero-shot  vanilla baseline trained with only labeled source-language data. This demonstrates the effectiveness of our method, achieved by enhancing the model's robustness towards perturbations on both labeled and unlabeled data. 

There is a considerable performance gap between ConNER and the translate-train methods (i.e., the upper section of Table~\ref{tab:main_result}). As translate-train suffers from unnatural translation and inaccurate label mapping, their performance is sub-optimal under full dataset settings where the vanilla baselines already present decent zero-shot performance.
This also shows the importance of exploiting unlabeled target-language data, apart from leveraging the available source training data and its parallel translation. 
Our ConNER also outperforms various state-of-the-art knowledge distillation methods (i.e., the middle section of Table~\ref{tab:main_result}). 
Soft-labels in knowledge distillation are often noisy, and such noise is likely to be amplified by forcing the student model's predictions to match the soft-labels. In comparison, our training method is more fault-tolerant, and induces better alignment between languages by training on parallel unlabeled data.
Furthermore, we compare ConNER with two consistency training baselines, which use word replacement~\citep{lowell2020unsupervised} or Gaussian noise~\citep{zheng2021consistency} as perturbations  (i.e., the lower section of Table~\ref{tab:main_result}). 

As we can see, word replacement might change the NER label of the original word and violate the consistency hypothesis, and thus lead to performance drops on some languages. On the other hand, Gaussian noise only applies perturbations with small magnitudes, and brings in limited improvement. In contrast, our ConNER leverages translation and dropout as high-quality perturbations. As a result, ConNER shows significant improvements over the consistency training baselines. 
Note that although TOF~\citep{zhang2021target} uses extra labeled MRC data during training, our ConNER still surpasses it on all of the target languages.


In general, our translation-based consistency training effectively utilizes unlabeled target data and encourages the model to extract language-invariant features,
while our dropout-based consistency further alleviates overfitting on source-language data for better cross-lingual adaptability.

\subsection{Low-resource NER}

We also investigate the effectiveness of ConNER on low-resource cross-lingual NER, where source-language training data is also limited. 
Specifically, we randomly sample 5\%, 10\% and 25\% examples from the original English training set as our low-resource training sets. We apply ConNER and evaluate on the original target-language test sets. As shown in Table~\ref{tab:low-resource}, ConNER is also effective under low-resource settings, achieving average performance improvements of 4.32, 2.20, and 1.77 F1 score compared with vanilla baseline respectively. We observe larger gains as the low-resource training set becomes smaller, which demonstrates the importance of leveraging unlabeled data and preventing overfitting under low-resource settings. 
Another interesting finding is that, using only 25\% of the original training data, our ConNER is already comparable to a vanilla NER model trained on the full training set.

\begin{table}[t!]
    \centering
    \resizebox{0.99\columnwidth}{!}{
    \begin{tabular}{clcccc}
    \hline\hline
         \#Train & \textbf{Method} & \textbf{De} & \textbf{Es} & \textbf{Nl} & \textbf{Avg} \\
         \hline
         \multirow{2}{*}{5\%} & vanilla &64.91	&65.76	&71.22	&67.30\\
         & ConNER & 69.94	&70.15	&74.76	&71.62\\
         \hline
         \multirow{2}{*}{10\%} & vanilla & 67.58    &69.92	&77.70	&71.73\\
         & ConNER & 70.88	&71.23	&79.70	&73.94\\
         \hline
         \multirow{2}{*}{25\%} & vanilla & 70.72 &	77.39 &	79.24 &	75.78\\
         & ConNER & 73.47 &	77.92 &	80.66 &	77.35\\

    \hline\hline
         
    \end{tabular}
    }
    \caption{Low-resource NER results.}
    \label{tab:low-resource}
\end{table}

\subsection{Distant Languages}

To evaluate the robustness of ConNER on a wider range of languages, we conduct experiments on three non-western target languages: Chinese (Zh), Arabic (Ar) and Hindi (Hi), from the WikiAnn dataset~\citep{pan-etal-2017-cross}.

\begin{table}[t!]
    \centering
    \resizebox{0.99\columnwidth}{!}{
    \begin{tabular}{lcccc}
    \hline\hline
         \textbf{Method} & \textbf{Zh} & \textbf{Ar} & \textbf{Hi} \\
         \hline
         vanilla                             & 33.10 & 53.00 & 73.00  \\
        ~\citet{wu2020explicit}              & 43.90 & 45.50 & 66.60  \\
        ~\citet{wu-etal-2020-single}         & 31.14 & 50.91 & 72.48  \\
         ConNER \textit{(Ours)}              & 39.17 & 59.62 & 74.49  \\
    \hline\hline
    \end{tabular}
    }
    \caption{Results on distant transfer pairs.}
    \label{tab:distant}
\end{table}

As shown in Table~\ref{tab:distant}, we observe the vanilla baseline has relatively low performance on these non-western languages, due to larger linguistic gaps. This might also introduce some noise in the choice of candidate spans for our method. 

Nonetheless, ConNER is still effective and demonstrates substantial improvements over the baseline. Such results verify that ConNER is general and robust for not only transfer pairs 
of the same language but also distant transfer pairs.

\section{Analysis}

\subsection{Ablation Study}

To analyze the contribution of each component and justify the framework design of ConNER, we conduct the following ablation studies: (1) \textbf{trans-unlabel}, where we only keep the translation-based consistency on unlabeled target data; (2) \textbf{dropout-label}, where we only keep the dropout-based consistency on labeled source data; (3) \textbf{trans-label}, where we translate labeled source language data to the target language, and calculate consistency loss between entities and their translations in the translated sentences. (4) \textbf{dropout-unlabel}, where we feed the same unlabeled target data twice through the encoder, and calculate consistency loss between the two output distributions of the same token.

\begin{table}[t]
    \centering
    \resizebox{0.99\columnwidth}{!}{
    \begin{tabular}{lcccc}
    \hline\hline
         \textbf{Method} & \textbf{De} & \textbf{Es} & \textbf{Nl} & \textbf{Avg} \\
         \hline
         vanilla & 72.10	& 79.64 & 81.29 & 77.68\\
         ConNER & 77.14	& 80.50	& 83.23	& 80.29\\
         \hline
         trans-unlabel & 76.87	& 80.76	& 81.66	& 79.77\\
         dropout-label & 74.46	& 80.90	& 81.63	& 78.99\\
         trans-label & 71.16	& 79.51	& 81.03	& 77.23\\
         dropout-unlabel & 61.72 & 77.80 &	80.80 &	73.44\\
    \hline\hline
    \end{tabular}
    }
    \caption{Ablation study.}
    \label{tab:ablation}
\end{table}

As shown in Table~\ref{tab:ablation}, both trans-unlabel and dropout-label achieve improved performance compared to the vanilla baseline. Nevertheless, the average improvement from trans-unlabel is more significant, demonstrating the importance of leveraging unlabeled target-language data for cross-lingual tasks. Combining trans-unlabel and dropout-label, our ConNER achieves further performance gains. 

However, when applying translation-based consistency to labeled data (trans-label) instead, we observe some performance drop. 
This is possibly because label information is already present, such that model trained on labeled source data already learns to make accurate predictions on its parallel translated sentence, and translation-based consistency becomes redundant.

Moreover, if we apply the dropout-based consistency on unlabeled target-language data (dropout-unlabel), we observe an unreasonably low performance (e.g., German). We attribute this to catastrophic forgetting of multilinguality. Without the guidance from the labels on target-language data, the model is likely to enter a failure mode that trivially minimizes the divergence for target-language tokens. As a result, target-language features are diverted away from the source-language in the representation space, causing the model to fail on the target language. In contrast, the translation-based consistency of ConNER makes use of parallel unlabeled data such that the source and target language are always aligned through the conjugate spans.


\subsection{Choice of Divergence Measure}

In this section, we investigate the effect of different divergence measures on the translation-based consistency. We consider three configurations: (1) KL-unlabel, where the predictions on the unlabeled target data are taken as ground-truth probabilities (Eqn.~\ref{eqn:kl1}); (2) KL-trans, where the predictions on the translated source data as ground-truth probabilities (Eqn.~\ref{eqn:kl2}); (3) bi-KL, which adopts a bidirectional KL divergence (Eqn.~\ref{eqn:kl3}).

\begin{subequations}
\small
\begin{gather}
    D_{\textrm{KL-unlabel}}= \textrm{KL}(P_{s}(z_{ij}|s_{ij})~||~P_{s}(z_{kl}|s'_{kl})) \label{eqn:kl1}\\
    D_{\textrm{KL-trans}}=\textrm{KL}(P_{s}(z_{kl}|s'_{kl})~||~P_{s}(z_{ij}|s_{ij})) \label{eqn:kl2}\\
\begin{aligned}
    D_{\textrm{bi-KL}}= \frac{1}{2}~[~&\textrm{KL}(P_{s}(z_{ij}|s_{ij})~||~P_{s}(z_{kl}|s'_{kl}))~ \\
+~&\textrm{KL}(P_{s}(z_{kl}|s'_{kl})~||~P_{s}(z_{ij}|s_{ij}))] \label{eqn:kl3}
\end{aligned}
\end{gather}
\end{subequations}

\begin{table}[t]
    \centering
    \resizebox{0.99\columnwidth}{!}{
    \begin{tabular}{lcccc}
    \hline\hline
         \textbf{Method} & \textbf{De} & \textbf{Es} & \textbf{Nl} & \textbf{Avg} \\
         \hline
         KL-unlabel & 76.08 &	80.55 &	80.06 &	78.90\\
         KL-trans & 76.65 &	80.31 &	81.22 &	79.39 \\
         bi-KL & 76.87 &	80.76 &	81.66 &	79.77\\
    \hline\hline
    \end{tabular}
    }
    \caption{Choice of divergence measure.}
    \label{tab:div}
\end{table}

From the experimental results in Table~\ref{tab:div}, we observe the overall performance of KL-trans is higher than KL-unlabel. This is probably due to the higher performance of the model on English data. 

By aligning predictions on unlabeled target-language data to the predictions on translated data in English, we propagate more accurate training signals from English to the target language and achieve better cross-lingual adaptation. Meanwhile, KL-unlabel also helps to reduce overfitting to English by taking target-language data as a reference. Bi-KL takes advantage of both KL-trans and KL-unlabel and achieves the best averaged performance.

\subsection{Case Study}

\begin{figure*}[t]
    \centering
    \hspace*{2cm}\includegraphics[width=0.93\textwidth]{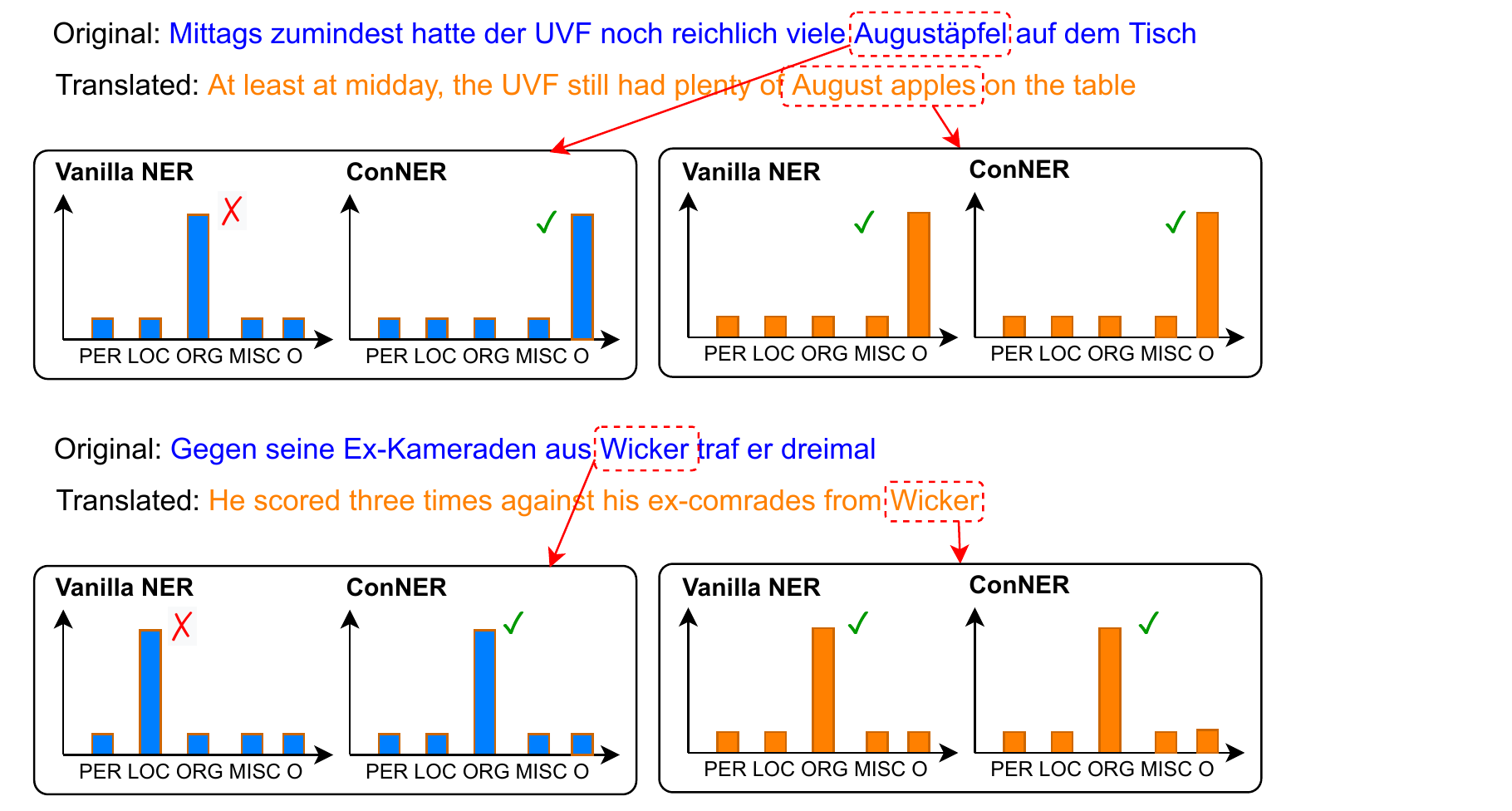}
    \caption{Case study on translation-based consistency.}
    \label{fig:case}
\end{figure*}

We conduct a case study to qualitatively show how translation-based consistency training benefits cross-lingual NER. As depicted in Fig.~\ref{fig:case}, the vanilla model predicts German word "\textit{Augustäpfel}" as an organization, but gives an inconsistent prediction for its English translation "\textit{August apples}" (i.e., O). In contrast, our model, which is enhanced by translation-based consistency training, attempts to behave similarly on the aligned spans and therefore is capable to calibrate the prediction on "\textit{Augustäpfel}". Similarly, while the vanilla model predicts German "\textit{Wicker}" as a location, our method leverages the context information to make the correct prediction as an organization.

\subsection{Robustness to Translation Systems}

In above experiments, we used Google Translate as our translation engine. To demonstrate the robustness of ConNER towards different machine translation systems, we rerun the experiments using Opus-MT~\citep{TiedemannThottingal:EAMT2020}, a popular open-source machine translation system. As shown in Table~\ref{tab:mt_sys}, there exists a minor performance drop when changing Google Translate to Opus-MT, possibly due to its lower translation quality. Nevertheless, it is observed that ConNER with Opus-MT is still superior to all baseline methods.

\begin{table}[h]
    \centering
    \resizebox{0.99\columnwidth}{!}{
    \begin{tabular}{lcccc}
    \hline\hline
         \textbf{Method} & \textbf{De} & \textbf{Es} & \textbf{Nl} & \textbf{Avg} \\
         \hline
         Google Translate   & 77.14 & 80.50 & 83.23 & 80.29 \\
         Opus-MT            & 75.86 & 80.60 & 82.47 & 79.64 \\
    \hline\hline
    \end{tabular}
    }
    \caption{Results using Google Translate and Opus-MT.}
    \label{tab:mt_sys}
\end{table}

\section{Related Work}

NER is the task to locate and classify named entities into pre-defined categories~\citep{xu2021better,zhou-etal-2022-melm,xu2022peerda}. Similar to other tasks under cross-lingual settings~\citep{zhang-etal-2021-cross,liu-etal-2022-enhancing,liu-etal-2022-towards-multi,zhou2022enhancing}, cross-lingual NER suffers from linguistic gaps.
Recent works on cross-lingual NER can be roughly categorized into three categories: (1) translate-train (2) knowledge distillation (3) consistency training.

\paragraph{Translate-train} creates target-language pseudo-labeled data by translating the source-language training set and projecting the NER labels of source-language texts to the translated texts.~\citet{xie2018neural} use bilingual embeddings to translate training data word-by-word and leverage self-attention for word order robustness.~\citet{bari2020zero} propose unsupervised translation using word-level adversarial learning and augmented fine-tuning with parameter sharing and feature augmentation.~\citet{li2020unsupervised} propose a warm-up mechanism to distill multilingual task-specific knowledge from the translated data in each language. Translate-train focuses on translating the available labeled source-language data, but it ignores the abundant unlabeled target-language data.

\paragraph{Knowledge distillation} trains the student model using the soft-labels obtained from a teacher model.~\citet{wu-etal-2020-single} apply knowledge distillation and propose a similarity measuring method to better weight the supervision from different teacher models.~\citet{ijcai2020-543} leverage both soft-labels and hard-labels from teacher models trained on the source data and translated training data.~\citet{chen-etal-2021-advpicker} use a discriminator to select less language-dependent target-language data via similarity to the source language.~\citet{zhang2021target} use extra machine reading comprehension data to enable multi-task adaptation and achieve improved NER performance. In general, knowledge distillation is sensitive to the noise in pseudo-label, especially when transferring to the target language.

\paragraph{Consistency training} applies perturbation on data or hidden states and regularizes the model to output consistent predictions. Existing works apply word replacement~\citep{lowell2020unsupervised} or Gaussian noise~\citep{zheng2021consistency} as perturbations. However, the validity and diversity of these noise or augmentations largely limit their performance~\cite{xie2020unsupervised}.~\citet{wang2021unsupervised} use back-translation as a high-quality augmentation. However, due to word alignment issues, they focus on the entity-appearance consistency in the whole sentence, but ignore the location of entities. 

\section{Conclusions}

In this work, we proposed a novel consistency framework for cross-lingual NER. Our method ConNER enhances the model's robustness towards perturbations on labeled and unlabeled data, via translation-based and dropout-based consistency. With translation-based consistency, we tackle the challenges of word alignment and token number changes via alignment-free projection and token-level to span-level conversion. Compared with multiple baseline methods, ConNER achieves consistent performance improvements.

\section{Limitations}

The method proposed requires the use of machine translation models or systems, which in some cases might not be easily accessible. Also, our translation-based consistency works the best when the selected candidate spans are likely to be entities and do not cross entity boundaries. Therefore, the strategy for candidate span selection could be a direction for future improvement.

\section*{Acknowledgements}
\label{sec:acknowledgements}

This research is partly supported by the Alibaba-NTU Singapore Joint Research Institute, Nanyang Technological University.

\bibliography{anthology}
\bibliographystyle{acl_natbib}




\end{document}